\documentclass[conference]{IEEEtran}
\IEEEoverridecommandlockouts
\usepackage{cite}
\usepackage{amsmath,amssymb,amsfonts}
\usepackage{algorithmic}
\usepackage{graphicx}
\usepackage{subfig}
\usepackage{textcomp}
\usepackage{xcolor}
\usepackage[hidelinks]{hyperref}
\def\BibTeX{{\rm B\kern-.05em{\sc i\kern-.025em b}\kern-.08em
    T\kern-.1667em\lower.7ex\hbox{E}\kern-.125emX}}

\begin{document}

\title{Benchmarking Hybrid Deep Learning Architectures for Predictive Maintenance in Industry 4.0\\
\thanks{\normalfont{This project was funded by Liveline Technologies. Liveline Technologies is dedicated to improving manufacturing performance by harnessing the power of Artificial Intelligence to automate complex processes and predict future problems.}}
\thanks{\normalfont{This is the author's version of this work. It is posted here for personal use, not for redistribution. The definitive version was published in the 2026 IEEE International Conference on AI and Data Analytics (ICAD), DOI: \href{https://ieeexplore.ieee.org/document/11608648}{10.1109/ICAD69378.2026.11608648}.}}
\thanks{\normalfont{\copyright~2026 IEEE. Personal use of this material is permitted. Permission from IEEE must be obtained for all other uses, in any current or future media, including reprinting/republishing this material for advertising or promotional purposes, creating new collective works, for resale or redistribution to servers or lists, or reuse of any copyrighted component of this work in other works.}}
}

\author{\IEEEauthorblockN{Zhengyang ``Cissy" Gu* \S}
\IEEEauthorblockA{\textit{Data Science, Liveline Technologies} \\
Livonia, MI, USA \\
cissy.gu@liveline.tech \\}
~\\

\and
\IEEEauthorblockN{Joseph E. Hernandez \ddag}
\IEEEauthorblockA{\textit{Engineering, Liveline Technologies} \\
Livonia, MI, USA \\
joseph.hernandez@liveline.tech }
~\\

\and
\IEEEauthorblockN{Thomas Cook}
\IEEEauthorblockA{\textit{Engineering, Liveline Technologies} \\
Livonia, MI, USA \\
thomas.cook@liveline.tech}
~\\

\and
\IEEEauthorblockN{John Burtenshaw}
\IEEEauthorblockA{\textit{Engineering, Liveline Technologies} \\
Livonia, MI, USA \\
john.burtenshaw@liveline.tech}
~\\

\and
\IEEEauthorblockN{Sean Scott}
\IEEEauthorblockA{\textit{Solutions, Liveline Technologies} \\
Livonia, MI, USA \\
sean.scott@liveline.tech}
~\\

\and
\IEEEauthorblockN{Chris Couch}
\IEEEauthorblockA{\textit{Chief Executive Officer, Liveline Technologies} \\
Livonia, MI, USA \\
chris.couch@liveline.tech}
}

\maketitle

\begin{abstract}
Predictive maintenance in Industry 4.0 refers to using data from sensors, machines, and production systems to estimate when equipment is likely to fail, so maintenance can be planned before a breakdown occurs \cite{b1}. However, a model that predicts maintenance may work perfectly in the lab but fail unexpectedly when applied to real factory data \cite{b2}.

To solve this ``reliability" gap, we evaluated six deep learning architectures across more than 700 experimental runs. 

We focused on the two dominant approaches in the field: Recurrent Neural Networks (RNNs), which process data step-by-step, like reading a sentence \cite{b3}, and Transformers, a recent dominant approach, which look at the entire sequence at once to spot important connections \cite{b4}. We examined whether Transformers still outperform recurrent neural networks (RNNs) when the data includes noise \cite{b5}.

We found that while Transformers excelled at tracking stable, slow-moving processes, they tend to overreact to chaotic data, mistakenly taking sensor noise for meaningful signals \cite{b6}. We also found that the hybrid method that combines a Long Short-Term Memory (LSTM) layer with a Transformer layer is more resilient to noisy data from factory shops \cite{b7}. Functioning as a noise filter, the LSTM smooths out data volatility, allowing the Transformer to focus on the bigger picture without being distracted \cite{b8}. The hybrid model did not just improve accuracy; it proved to be significantly more consistent than complex models, delivering reliable predictions regardless of how chaotic the underlying system became.

\end{abstract}

\begin{IEEEkeywords}
Industrial AI, Predictive Maintenance, Deep Learning, Transformers, LSTM, Reliability, Industry 4.0
\end{IEEEkeywords}

\section{Introduction}
Modern manufacturing has shifted from reactive repairs to data-driven predictive maintenance enabled by the Industrial Internet of Things (IIoT) \cite{b9}. While traditional statistical methods like ARIMA have served the industry well for linear systems \cite{b10}, they often struggle to model the non-linear complexity of high-frequency IoT data \cite{b11}. Consequently, the field has increasingly adopted Deep Learning: utilizing Convolutional Neural Networks (CNNs) for local feature extraction \cite{b12}, Long Short-Term Memory (LSTM) networks for temporal tracking \cite{b3}, and recently, Transformers for capturing long-range dependencies \cite{b4}.

However, a critical gap remains in the literature: \textit{Stability} \cite{b13}. In academic research, a model that achieves state-of-the-art performance on a single ``lucky'' training run is often considered a success \cite{b14}. On the factory floor, a model that works once but fails to converge next time is a problem. Many studies ignore this volatility and focus on peak scores, which can hide how unstable pure Attention models can be in noisy industrial settings \cite{b6}.

We address this reliability gap by conducting a ``stress test'' of six deep learning architectures. Instead of reporting a single best run, we ran over 700 independent training sessions to quantify each model's variance, ruling out the influence of random start weights \cite{b15}. We examine whether mixing specific layer types, pairing the local feature extraction of \textbf{Convolutions} \cite{b16} or the sequential filtering of \textbf{Recurrence} with the global context of \textbf{Attention}, can produce a model that performs reliably in statistical tests, rather than only on rare, lucky runs.

Our contributions are:
\begin{itemize}
    \item \textbf{Synthetic Stress Testing:} We evaluate models against distinct synthetic regimes: High-Inertia (stable) and Chaotic (stochastic), to understand their divergent behaviors under contrasting simulated industrial conditions \cite{b17}.
    \item \textbf{The LSTM-Transformer Solution:} We identify the LSTM-Transformer as the most suitable architecture. We demonstrate that using an LSTM as a ``noise filter'' first allows the Transformer to capture global trends without overfitting to local chaos \cite{b7}.
    \item \textbf{Quantifying Reliability:} By analyzing model performance distributions, we challenge the idea that ``bigger is better" \cite{b18}. We show that the complex Tri-Hybrid model suffers from diminishing returns, proving that a targeted design approach yields better results than blindly adding complexity.
\end{itemize}

\section{Methodology}

\subsection{Data Acquisition and Synthetic Regime Design}
We used two contrasting datasets to test whether our Deep Learning architectures are resilient across different synthetic regimes. The first dataset represents a ``High-Inertia” process, characterized by slow, predictable changes, while the second represents a ``Chaotic” process, characterized by high stochasticity \cite{b19}.

\subsubsection{Dataset 1: Smart Manufacturing (High-Inertia)}
For the stable regime, we utilized the Smart Manufacturing Process Data from Kaggle \cite{b20}. This dataset contains 10,000 time-stamped observations collected at one-minute intervals from an industrial machine. It captures seven key operational features, including Temperature ($^\circ$C), Machine Speed (RPM), Production Quality Score, Vibration Level (mm/s), and Energy Consumption (kWh) \cite{b20}. 

\textbf{Synthetic Target Injection:} To create a target variable ($Y$) that reflects realistic equipment degradation, we injected a ``Penalty'' relationship into the raw sensor data (Eq. \ref{eq:manuf_physics}) and applied Z-score normalization to standardize the inputs \cite{b21}.

\begin{equation}
E_t = 100 - 2Z_{T} - 3Z_{V} - 1|Z_{R}| - 1.5Z_{E} + \mathcal{N}(0, 0.5)
\label{eq:manuf_physics}
\end{equation}

Where:
\begin{itemize}
    \item \textbf{$E_t$ (Equipment Health):} The calculated efficiency score at time $t$.
    \item \textbf{$Z_{T}$ (Temperature):} The temperature. A higher value indicates overheating, which we penalize with a significant coefficient (-2).
    \item \textbf{$Z_{V}$ (Vibration):} The vibration. This factor carries the heaviest penalty (-3), as high vibration is typically the leading indicator of mechanical instability \cite{b1}.
    \item \textbf{$Z_{R}$ (RPM):} The absolute value of machine speed. We use the absolute value here because any deviation from the norm, whether under-speed or over-speed, is dangerous to machine health.
    \item \textbf{$Z_{E}$ (Energy):} The energy consumption. Abnormal spikes in power usage are treated as signs of inefficiency (-1.5).
\end{itemize}

The variable $\mathcal{N}(0, 0.5)$ adds slight Gaussian noise to simulate measurement error. In addition, we applied a rolling window ($t=3$) to these inputs before calculation. This introduces \textit{inertia}, meaning the system state at time $t$ is highly predictive of $t+1$, mimicking thermal or mechanical latency \cite{b10}.

It is important to note that because $E_t$, the equipment health score, is computed from a known deterministic formula with additive Gaussian noise, the learning task is fundamentally function approximation rather than prediction of organic equipment degradation. Our models may effectively be reverse-engineering this equation. While this is one of our limitations, this controlled setup is useful for isolating different sensor differences under known conditions.

\subsubsection{Dataset 2: Augmented IoT (Chaotic)}
For the chaotic regime, we utilized the Real-Time IoT Driven Production System dataset from Kaggle \cite{b22}. It encompasses diverse machine types and tracks key process parameters such as temperature, vibration, power consumption, and material flow rate, alongside production outcomes like error rates, downtime, and maintenance needs \cite{b22}.

\textbf{Synthetic Target Injection:} Although the original dataset contains only 4,000 samples, we expanded the dataset to 80,000 samples across four machine profiles (M001--M004) by applying distinct physical profiles to each machine. We use the term ``Augmented'' to reflect this simulation of a factory environment rather than a large-scale data collection from a real factory production. Specifically, for Machine M001, we injected a chaotic relationship driven by instantaneous error rates rather than smoothed trends (Eq. \ref{eq:iot_physics}).

\begin{equation}
E_{t, M001} = 100 - 10 \cdot \text{ErrorRate}_t - 2 \cdot \text{Vib}_t + \mathcal{N}(0, 2.0)
\label{eq:iot_physics}
\end{equation}

Where:
\begin{itemize}
    \item \textbf{$E_{t, M001}$:} The instantaneous efficiency score for Machine 001.
    \item \textbf{$\text{ErrorRate}_t$:} The real-time error rate, assigned with a massive penalty coefficient (-10). Unlike the smoothed metrics in Dataset 1, a spike in error rate here will cause an immediate, sharp drop in efficiency, simulating a sudden machine jam or malfunction.
    \item \textbf{$\text{Vib}_t$:} The vibration, assigned with a moderate penalty coefficient.
    \item \textbf{$\mathcal{N}(0, 2.0)$:} A high-variance Gaussian noise term. We set the standard deviation to $2.0$ (four times the noise in Dataset 1) to simulate the high-noise environment of a factory floor.
\end{itemize}

Although the injected Gaussian noise is a simplification, real factory disturbances are typically non-stationary and non-Gaussian. This formula results in a system with reduced inertial memory: the state at time $t$ exhibits weak correlation with $t+1$, providing a testbed for the stability of our deep learning architectures in high-volatility environments \cite{b17}.

\subsection{Comparative Synthetic Profiling}
We used three statistical metrics to compare the differences between our ``High-Inertia" and ``Chaotic" datasets.

\begin{figure}[htbp]
    \centering
    \begin{minipage}{0.48\columnwidth}
        \centering
        \includegraphics[width=\linewidth]{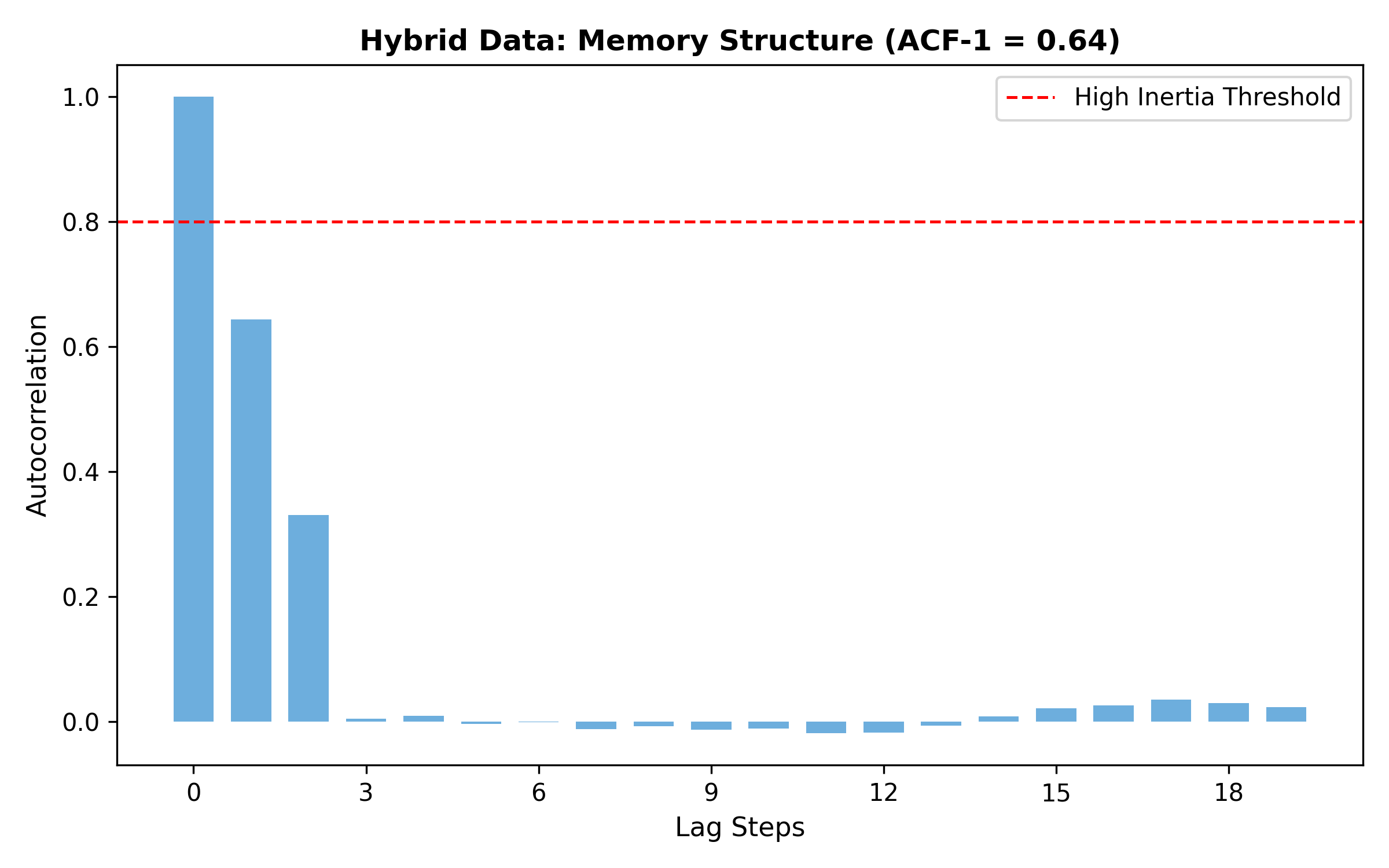}
        \par\vspace{2pt}
        \footnotesize \centering (a) Dataset 1: Manufacturing
    \end{minipage}\hfill
    \begin{minipage}{0.48\columnwidth}
        \centering
        \includegraphics[width=\linewidth]{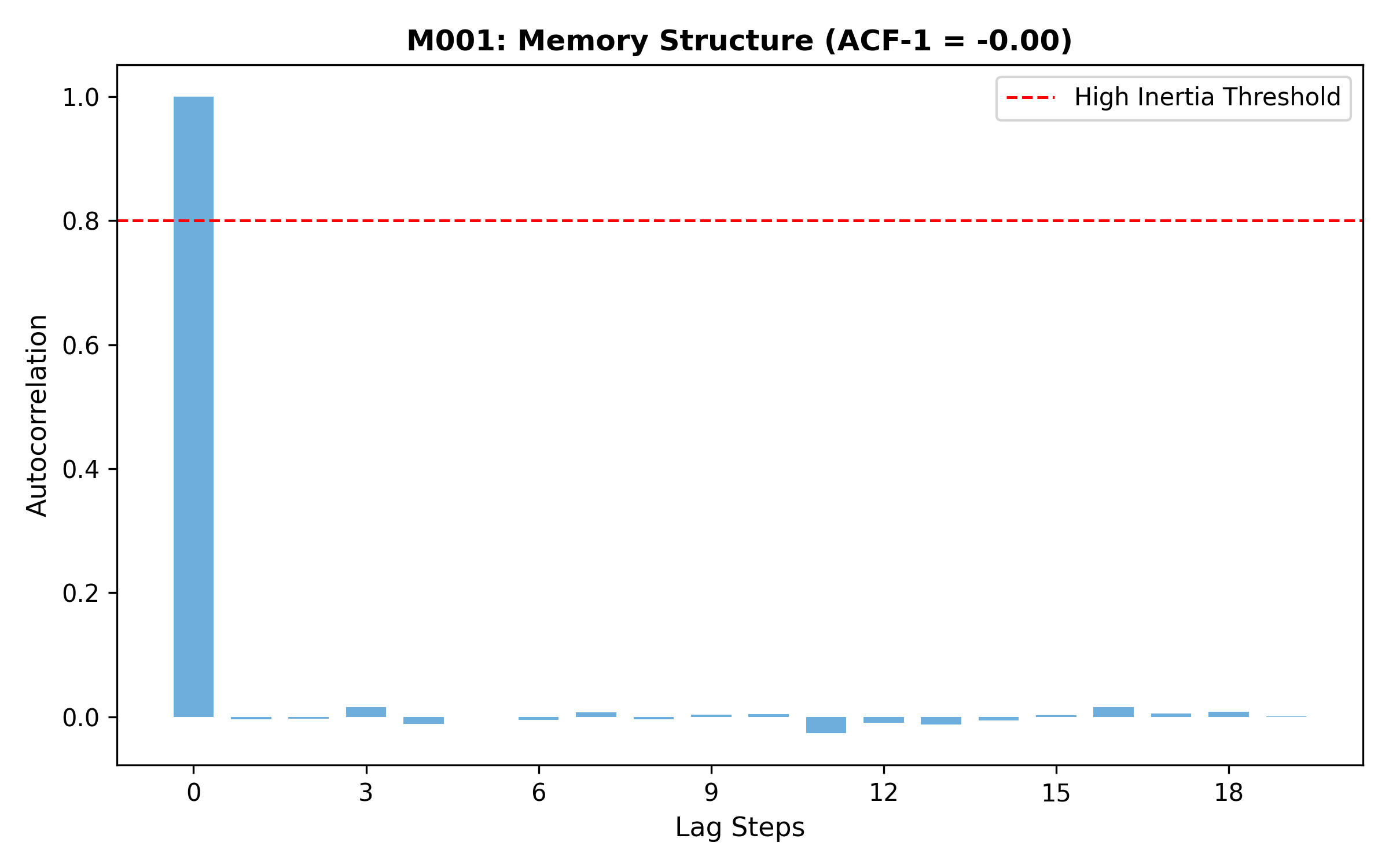}
        \par\vspace{2pt}
        \footnotesize \centering (b) Dataset 2: IoT
    \end{minipage}
    \caption{\textbf{ACF Comparison.} (a) Manufacturing data show high inertia. (b) IoT data show limited temporal dependence.}
    \label{fig:acf_comparison}
\end{figure}

\textbf{1. Memory Structure (ACF):} Fig. \ref{fig:acf_comparison} plots the Autocorrelation Functions (ACF). The Manufacturing data (a) exhibit \textit{strong-moderate temporal memory} (ACF $>$ 0.2) for 3 time steps, meaning the system's current state is a reliable predictor of its future. In contrast, the IoT data (b) show a sharp drop in autocorrelation. This indicates a system with \textit{weak temporal dependence}, where historical trends provide minimal guidance for future values \cite{b10}.

\begin{figure}[htbp]
    \centering
    \begin{minipage}{0.48\columnwidth}
        \centering
        \includegraphics[width=\linewidth]{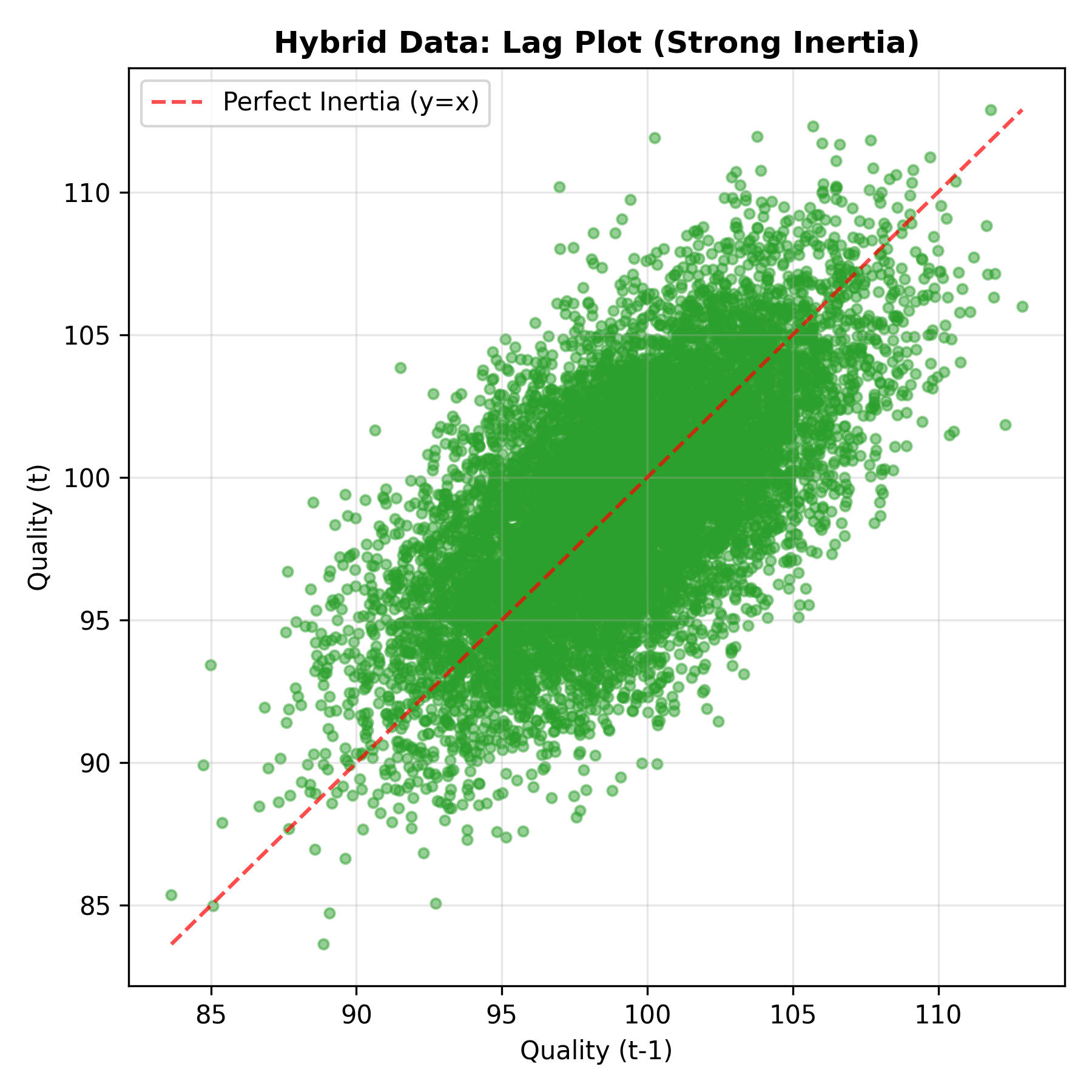}
        \par\vspace{2pt}
        \footnotesize \centering (a) Dataset 1: Manufacturing
    \end{minipage}\hfill
    \begin{minipage}{0.48\columnwidth}
        \centering
        \includegraphics[width=\linewidth]{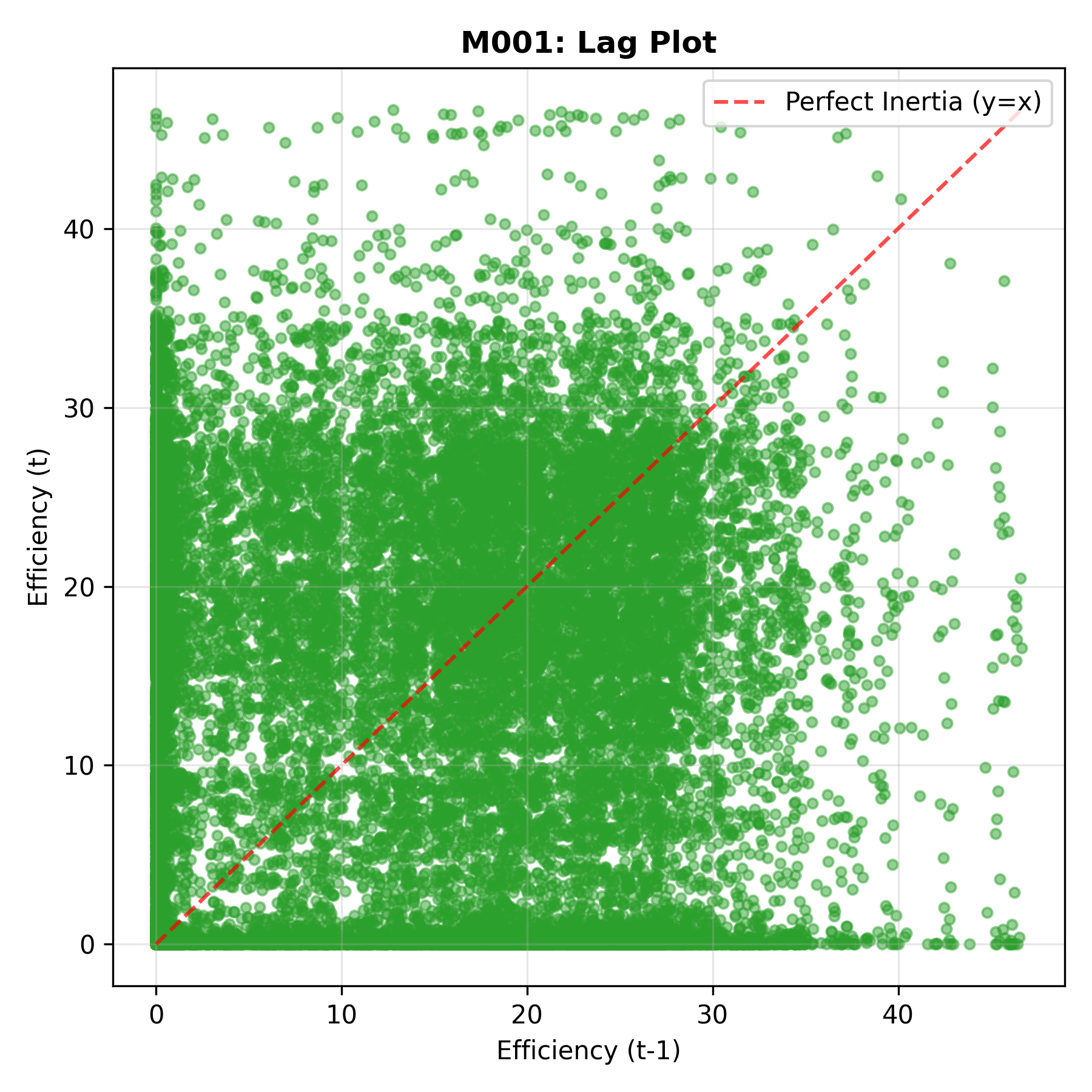}
        \par\vspace{2pt}
        \footnotesize \centering (b) Dataset 2: IoT
    \end{minipage}
    \caption{\textbf{Predictability Comparison.} (a): The tight diagonal relationship indicates a predictable system. (b): The scattered cloud indicates a noisy, chaotic system.}
    \label{fig:lag_comparison}
\end{figure}

\textbf{2. System Predictability (Lag Plots):} The Lag Plots in Fig. \ref{fig:lag_comparison} visualize the relationship between $y_t$ and $y_{t-1}$. The Manufacturing data cluster closely to a diagonal line, indicating that the system's physical relationship is far stronger than the sensor noise. By comparison, the IoT data appear scattered. This lack of a clear pattern confirms the system's high stochasticity, where the previous state ($y_{t-1}$) provides little predictive information about the current state ($y_t$).

\textbf{3. State Stability (Markov):} We discretized efficiency scores into Low/Avg/High categories to estimate transition probabilities between states using Markov Chains \cite{b23}. The Manufacturing system displays high state persistence; if it is in an Avg-efficiency state now, it is 83\% likely to remain there. 

The IoT system, conversely, reveals a distinct pattern. Rather than random switching, the matrix shows a strong tendency toward degradation. If it is in a High-efficiency state now, there is only a 5\% probability of remaining there, but a 95\% probability of an imminent drop to \textit{Avg} or \textit{Low} states. Even within the \textit{Low} state, the system switches between \textit{Low} (48\%) and \textit{Avg} (46\%), confirming that the M001 profile is chaotic and driven by error.

\begin{figure}[htbp]
    \centering
    \begin{minipage}{0.48\columnwidth}
        \centering
        \includegraphics[width=\linewidth]{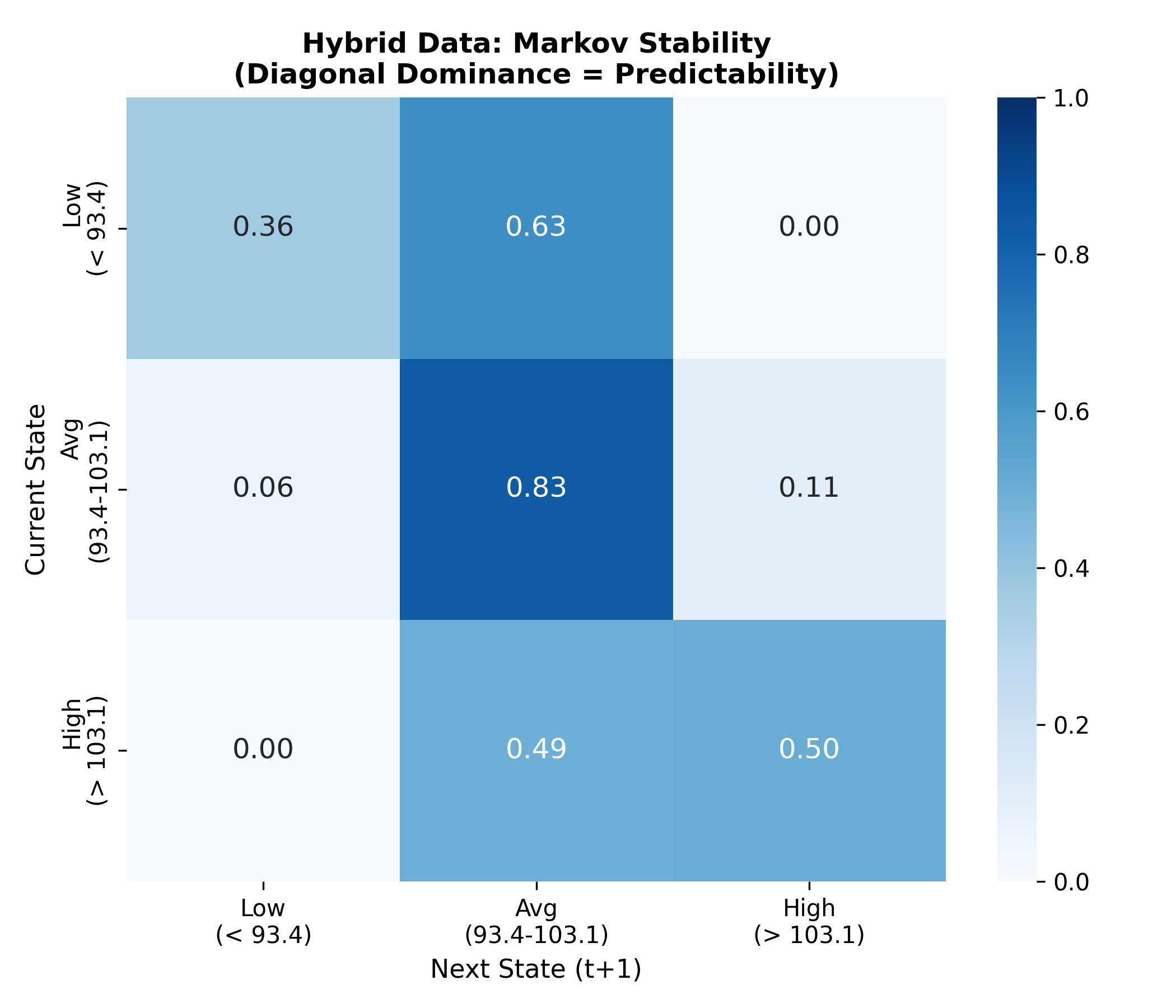}
        \par\vspace{2pt}
        \footnotesize \centering (a) Dataset 1: Manufacturing
    \end{minipage}\hfill
    \begin{minipage}{0.48\columnwidth}
        \centering
        \includegraphics[width=\linewidth]{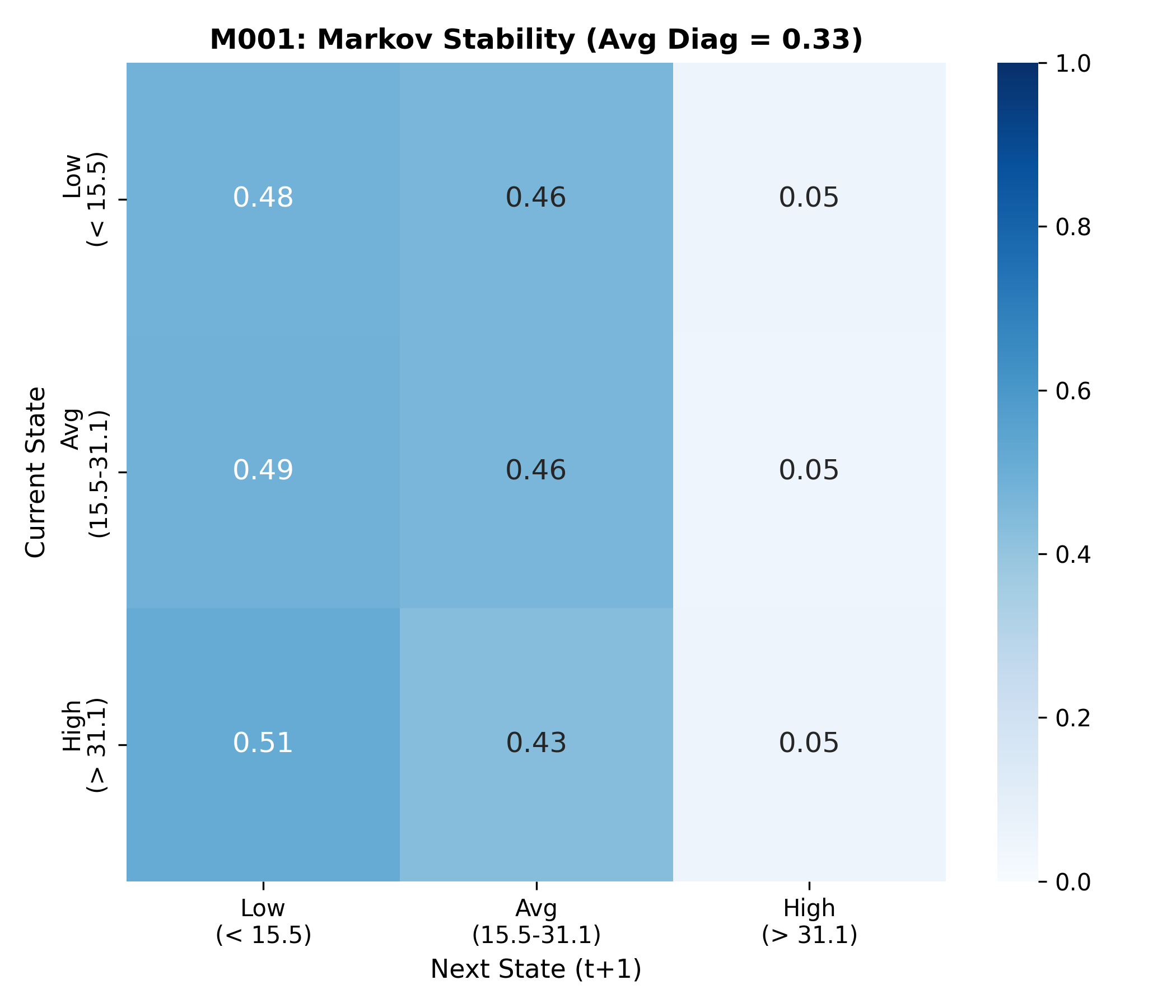}
        \par\vspace{2pt}
        \footnotesize \centering (b) Dataset 2: IoT
    \end{minipage}
    \caption{\textbf{Stability Comparison.} (a): Manufacturing data show high state persistence. (b): IoT data show state degradation.}
    \label{fig:markov_comparison}
\end{figure}

\subsection{Data Preprocessing}
We applied a consistent data preprocessing strategy to each dataset:
\begin{enumerate}
    \item \textbf{Window Size ($T$):} We applied a sliding window size based on the system's profile. For the Manufacturing data, we used $T=20$ steps to capture slow-moving trends. For the chaotic IoT data, we shortened this to $T=10$, forcing the models to focus on short-term volatility rather than distant history \cite{b24}.
    \item \textbf{Normalization:} We scaled all features to the [0, 1] range (using Min-Max Scaler) to ensure stable training \cite{b21}.
\end{enumerate}

\section{Experimental Setup}

We stress-tested our Deep Learning architectures over 700 individual training sessions. While many studies cherry-pick a single best run that favors the model, we repeated each model 20 times to capture its full performance distribution. This stress-test approach guarantees that our results reflect true reliability: how the model performs on average, rather than on a ``lucky seed'' \cite{b14}.

\subsection{Experimental Design}
We split the testing into two phases, using the two machine datasets described in Section II.

\subsubsection{Phase 1: High-Inertia Regime (Manufacturing Dataset)}
We first tested all six models on the stable Smart Manufacturing data.
\begin{itemize}
    \item \textbf{Experiments:} 20 independent runs per model, each initialized with a different random seed. That is, $6 \text{ Models} \times 20 \text{ Runs} = 120 \text{ Training Sessions}$.
    \item \textbf{Training Config:} We utilized the Adam optimizer \cite{b25} with a learning rate of $0.001$ and Mean Squared Error (MSE) loss. Training was capped at 20 epochs with a batch size of 32, and Early Stopping (patience=5) was employed to prevent overfitting \cite{b26}.
\end{itemize}

\subsubsection{Phase 2: Chaotic Regime (IoT Dataset)}
We performed separate experiments for each specific machine (M001, M002, M003, M004) and a fifth ``Global'' experiment combining all data.
\begin{itemize}
    \item \textbf{Experiments:} 20 independent runs per model, for 5 experiments. That is, $6 \text{ Models} \times 5 \text{ Subsets} \times 20 \text{ Runs} = 600 \text{ Training Sessions}$.
    \item \textbf{Training Config:} To accommodate the high stochasticity of the IoT data, we adjusted the hyperparameters: a reduced hidden dimension of 32 units (to prevent overfitting to noise), a larger batch size of 128 (to stabilize gradient updates) \cite{b27}, and L2 regularization ($\lambda=0.001$).
\end{itemize}

\subsection{Model Architectures}
We implemented six distinct deep learning architectures, ranging from standard baseline models to complex hybrids. 

\begin{figure*}[t!]
    \centering
    % (a) CNN
    \begin{minipage}[b]{0.16\textwidth}
        \centering
        \includegraphics[width=\linewidth]{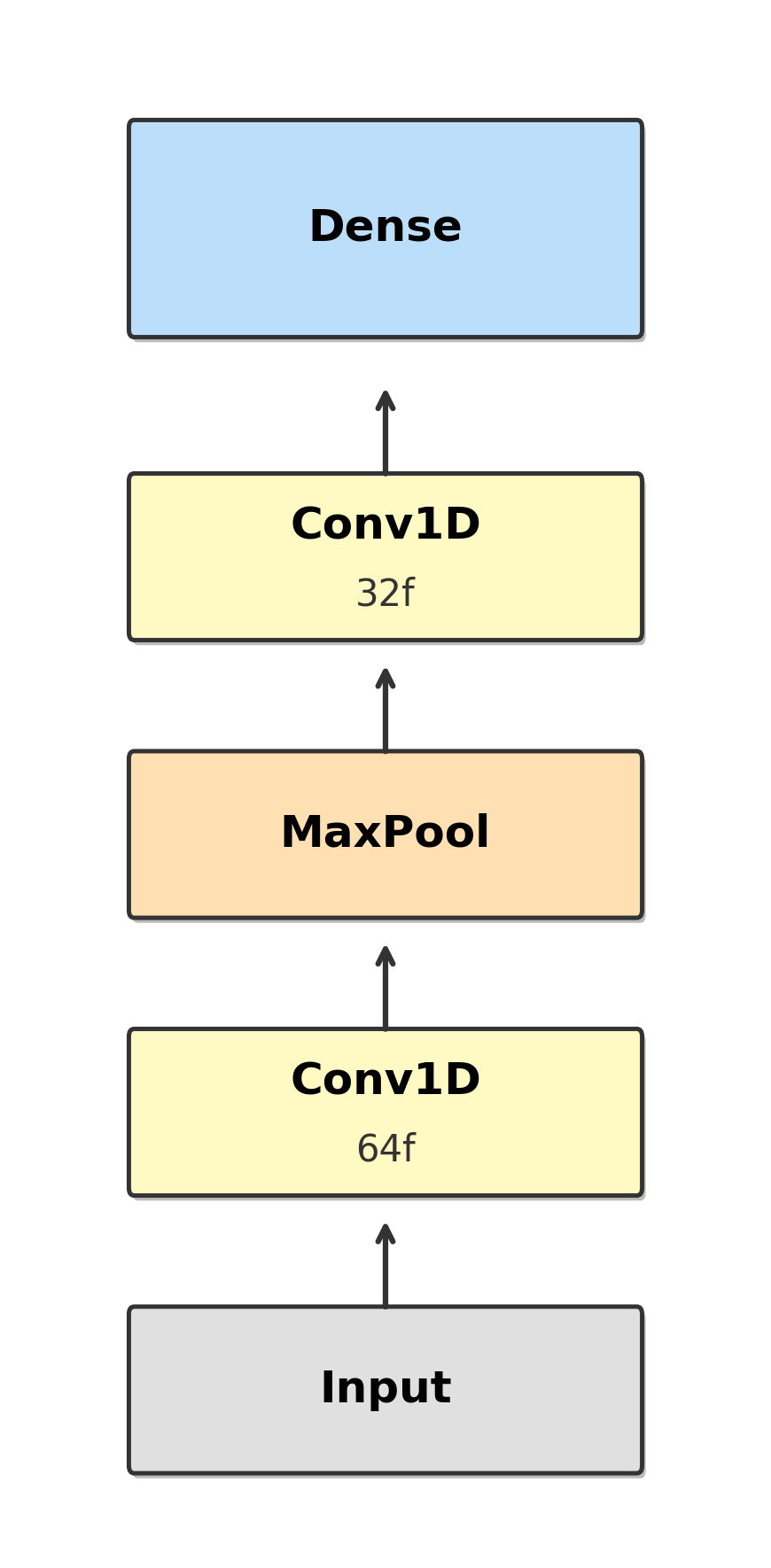} 
        \par\vspace{2pt}
        \footnotesize (a) CNN
    \end{minipage}
    \hfill % Rubber space for alignment
    % (b) LSTM
    \begin{minipage}[b]{0.16\textwidth}
        \centering
        \includegraphics[width=\linewidth]{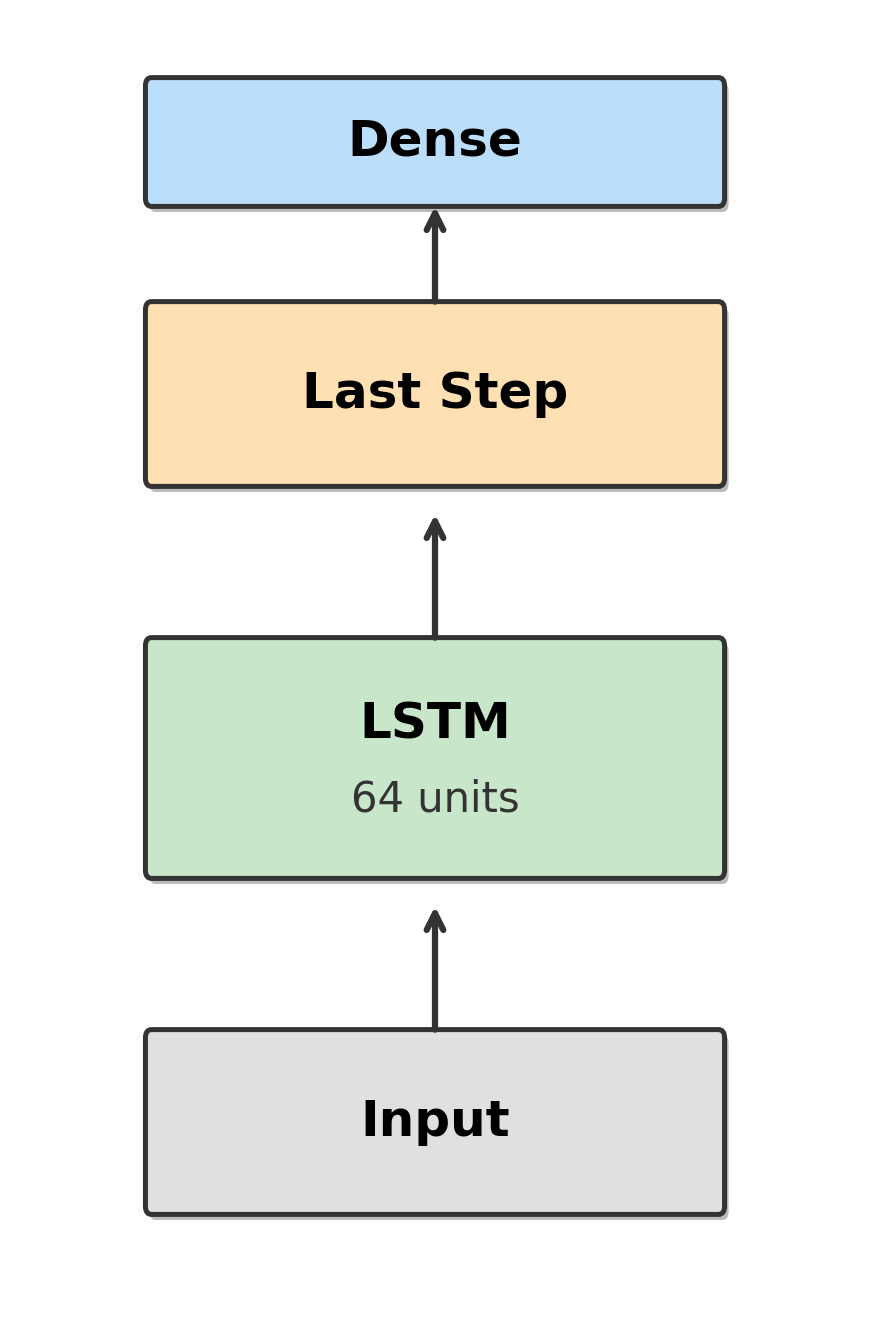}
        \par\vspace{2pt}
        \footnotesize (b) LSTM
    \end{minipage}
    \hfill
    % (c) CNN-LSTM
    \begin{minipage}[b]{0.16\textwidth}
        \centering
        \includegraphics[width=\linewidth]{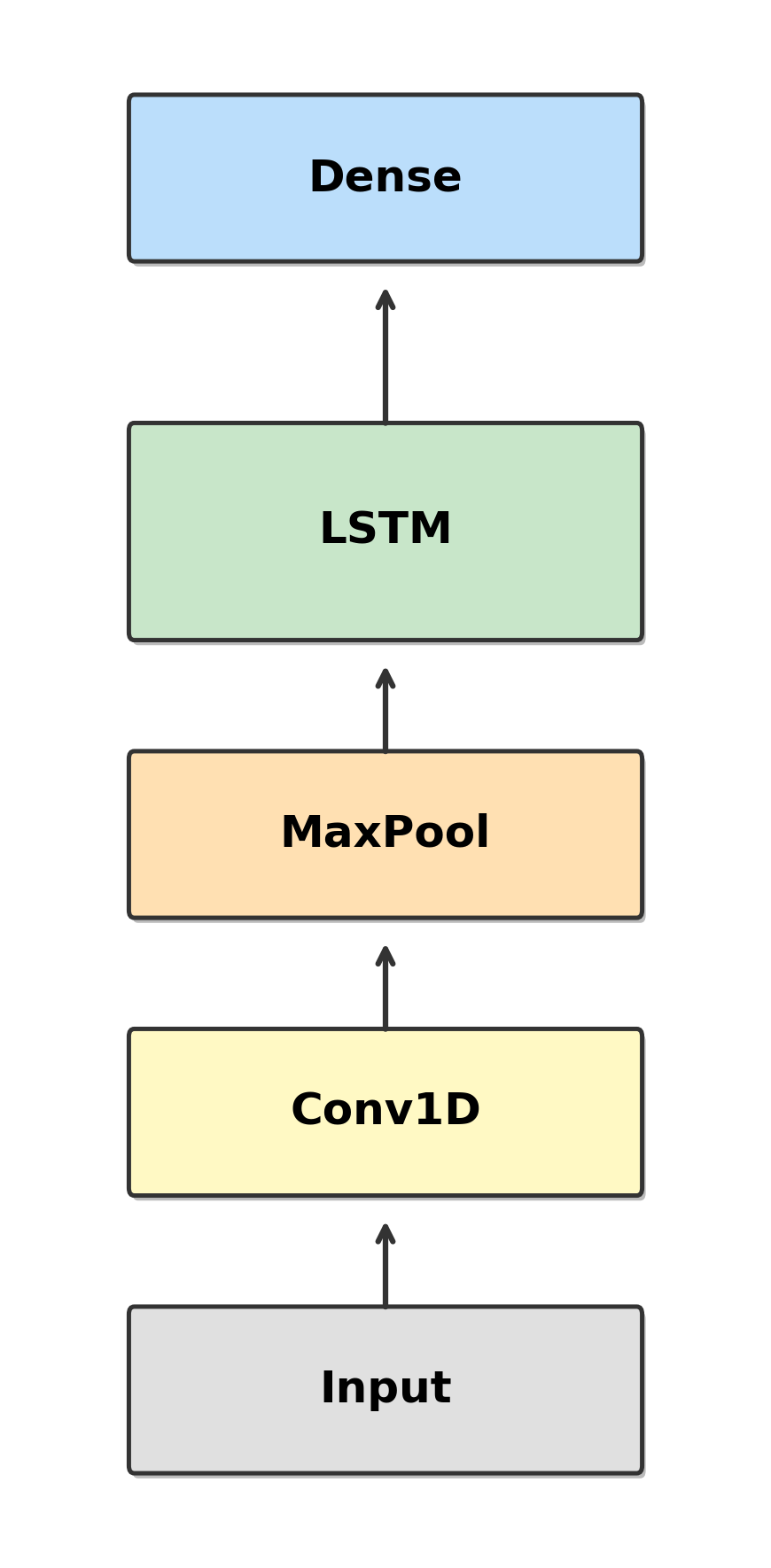}
        \par\vspace{2pt}
        \footnotesize (c) CNN-LSTM
    \end{minipage}
    \hfill
    % (d) Transformer
    \begin{minipage}[b]{0.16\textwidth}
        \centering
        \includegraphics[width=\linewidth]{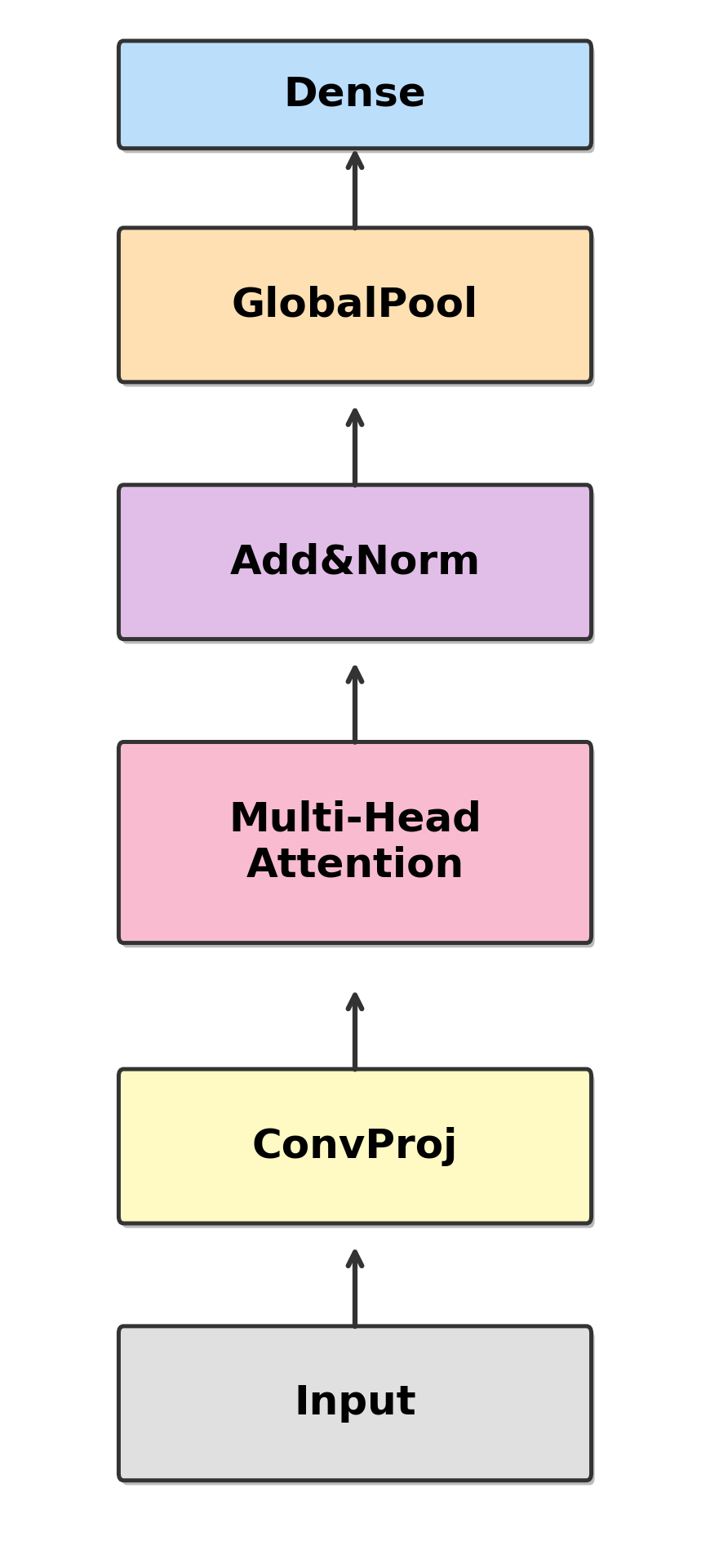}
        \par\vspace{2pt}
        \footnotesize (d) Transformer
    \end{minipage}
    \hfill
    % (e) LSTM-Trans
    \begin{minipage}[b]{0.16\textwidth}
        \centering
        \includegraphics[width=\linewidth]{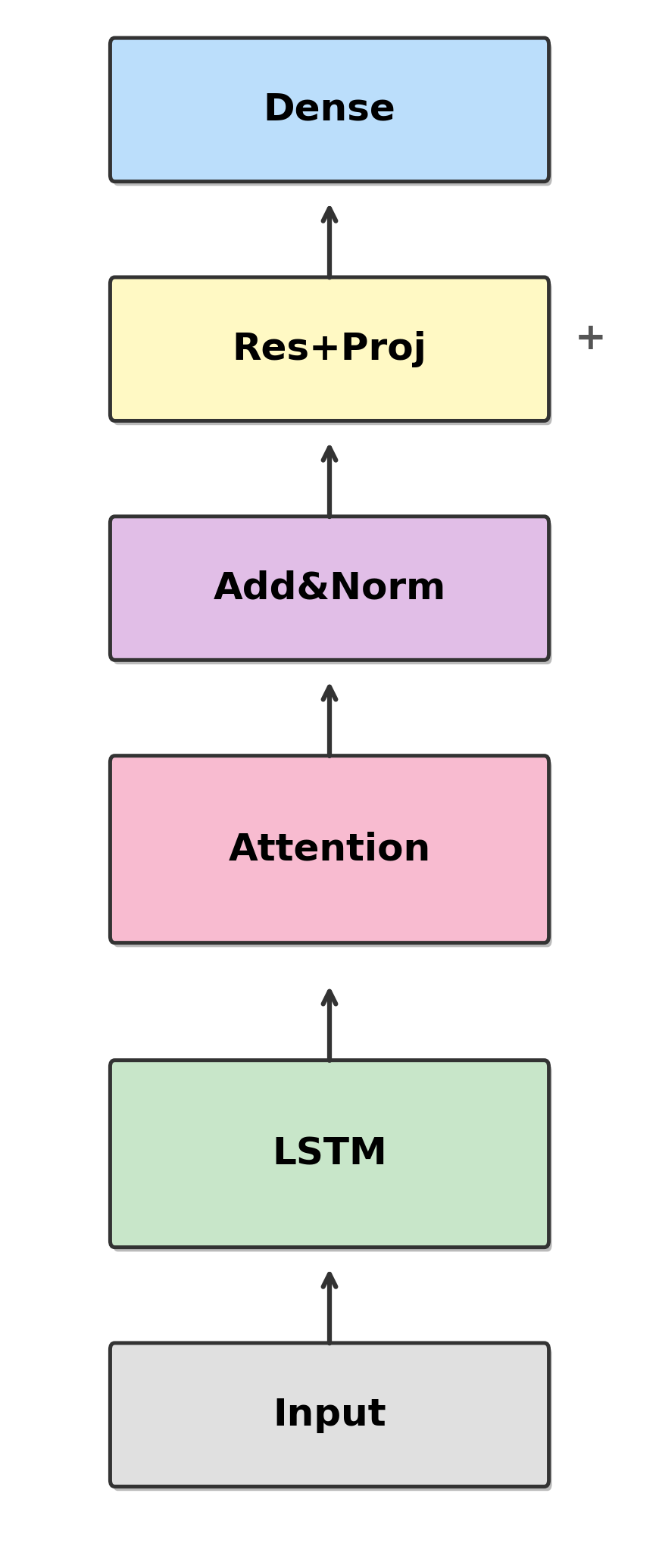}
        \par\vspace{2pt}
        \footnotesize (e) LSTM-Transformer
    \end{minipage}
    \hfill
    % (f) Tri-Hybrid
    \begin{minipage}[b]{0.16\textwidth}
        \centering
        \includegraphics[width=\linewidth]{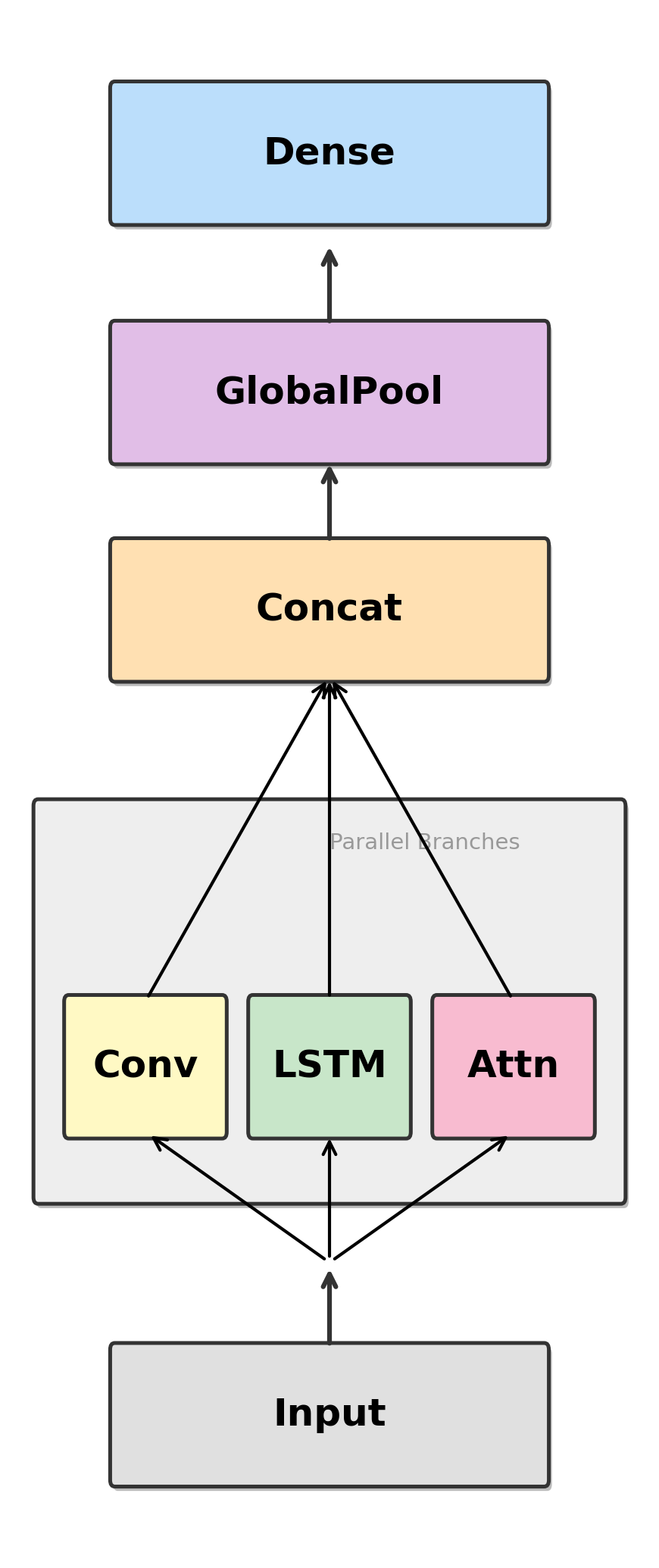}
        \par\vspace{2pt}
        \footnotesize (f) Tri-Hybrid
    \end{minipage}
    
    \vspace{0.2cm} % Space between images and main caption
    \caption{\textbf{Architecture Comparison.} Data flow (Bottom-to-Top) for the six evaluated models. (a) CNN (b) LSTM (c) CNN-LSTM (d) Pure Attention. (e) Our proposed hybrid (LSTM+Attention). (f) Parallel ``Wide \& Deep''.}
    \label{fig:model_architectures}
\end{figure*}

Fig. \ref{fig:model_architectures} provides a structural overview of the 6 models evaluated.

% --- FIGURE: 1 ROW of 6 COLUMNS ---

\subsubsection{Convolutional Neural Network (CNN / ResNet)}
We used Convolutions to detect local patterns rather than time-based sequences \cite{b12}.
\begin{itemize}
    \item \textbf{Setup:} For the stable manufacturing data, we used a standard stack of 1D Convolutions. For the chaotic IoT data, we added Residual connections (ResNet) to preserve the raw signal against heavy noise \cite{b18}, aided by Batch Normalization.
    \item \textbf{Assessment:} Since it lacks memory, it often misses the peak values in slow-moving systems.
\end{itemize}

\subsubsection{Long Short-Term Memory (LSTM)}
The standard baseline model for temporal sequence modeling \cite{b3}.
\begin{itemize}
    \item \textbf{Setup:} We applied a single LSTM layer ($64$ units for Manufacturing, $32$ for IoT). Uniquely for the IoT dataset, the final hidden state $h_T$ is concatenated with the raw input $x_T$ (Wide \& Deep) to allow the model to react to immediate shocks \cite{b28}.
    \item \textbf{Assessment:} The LSTM acts like a \textit{Low-Pass Filter} \cite{b29}. By processing data sequentially, it naturally suppresses high-frequency Gaussian noise. However, its sequential nature limits its ability to capture global dependencies compared to Attention mechanisms.
\end{itemize}

\subsubsection{CNN-LSTM}
A sequential hybrid to compress data before analyzing it.
\begin{itemize}
    \item \textbf{Setup:} This model uses a CNN first to compress noisy raw data into cleaner abstract features, which the LSTM then processes \cite{b30}.
    \item \textbf{Assessment:} This model offers a middle ground. While it filters noise well, aggressive downsampling by the CNN can sometimes discard important details.
\end{itemize}

\subsubsection{Transformer (Encoder-Only)}
This model uses Self-Attention to look at the entire history at once, rather than step-by-step (like LSTM) \cite{b3}.
\begin{itemize}
    \item \textbf{Setup:} We projected raw sensor data into a 64-dimensional latent space. The Self-Attention mechanism then computes a weighted sum of all time steps, allowing the model to ``attend'' to any point in history, in order to find global correlations \cite{b4}.
    \item \textbf{Assessment:} It dominates in stable environments where context is king. However, in chaotic regimes, the global attention mechanism tends to overfit to stochastic noise, resulting in higher variance and lower accuracy.
\end{itemize}

\subsubsection{LSTM-Transformer (Proposed)}
Our solution to the stability problem.
\begin{itemize}
    \item \textbf{Setup:} Unlike standard Transformers that project raw data linearly, we use an LSTM as the tokenizer. The LSTM processes the sequence step-by-step, filtering out local Gaussian noise and encoding short-term trends \cite{b8}. The hidden states are then passed to the Transformer layer.
    \item \textbf{Assessment:} It offers the \textit{best of both worlds}. The LSTM stabilizes the input (Noise Filtering), while the Transformer captures long-range dependencies (Global Context). This unique combination makes it the only ``Regime-Agnostic'' model in our benchmark \cite{b7}.
\end{itemize}

\subsubsection{Tri-Hybrid Architecture}
We combined everything (Conv + LSTM + Attention) to test the limits of complexity.
\begin{itemize}
    \item \textbf{Setup:} We built a massive pipeline to see if deeper models perform better. We employed a ``\textbf{Wide \& Deep}'' architecture. This mechanism concatenates the raw input at the final time step ($x_T$) with the deep network output before the final dense layer, preserving high-frequency signal details alongside deep features \cite{b28}.
    \item \textbf{Assessment:} It matched the LSTM-Transformer but failed to beat it. Its diminishing returns did not justify its structural complexity \cite{b18}. 
\end{itemize}

\section{Results and Discussion}

\subsection{Scenario A: The High-Inertia Regime (Smart Manufacturing)}

Table \ref{tab:manuf_results} summarizes the aggregated performance of the six Deep Learning architectures over 20 independent runs on the Smart Manufacturing dataset.

\begin{table}[htbp]
\caption{Manufacturing Dataset Results (20 Runs)}
\begin{center}
\begin{tabular}{|l|c|c|c|}
\hline
\textbf{Model} & \textbf{$R^2$ Score (Mean $\pm$ Std)} & \textbf{RMSE} & \textbf{MAE} \\
\hline
Transformer & \textbf{0.644 $\pm$ 0.002} & \textbf{2.392} & \textbf{1.938} \\
LSTM-Trans & \textbf{0.644 $\pm$ 0.002} & 2.393 & 1.943 \\
LSTM & 0.643 $\pm$ 0.002 & 2.396 & 1.943 \\
CNN-LSTM & 0.642 $\pm$ 0.003 & 2.399 & 1.946 \\
Tri-Hybrid & 0.640 $\pm$ 0.003 & 2.406 & 1.951 \\
CNN & 0.633 $\pm$ 0.005 & 2.428 & 1.967 \\
\hline
\end{tabular}
\label{tab:manuf_results}
\end{center}
\end{table}

\subsubsection{Conclusion on Predictive Power}
The results establish the \textit{Transformer} as the optimal architecture for high-inertia industrial processes:
\begin{itemize}
    \item \textbf{Transformer Dominance:} The pure \textbf{Transformer} achieved the best performance across all three metrics. It was tied for the highest $R^2$ ($0.644$) but achieved the lowest error rates (RMSE: $2.392$, MAE: $1.938$), demonstrating the most reliable capability to predict high-inertia processes \cite{b3}.
    \item \textbf{Baseline Underperformance:} The \textbf{CNN} consistently underperformed, exhibiting the highest error (RMSE: $2.428$, MAE: $1.967$) and variance ($\sigma=0.005$). This confirms that a standard convolution on its own is insufficient to capture continuous variation in the data without a dedicated memory mechanism \cite{b16}.
\end{itemize}

\subsubsection{Visual Analysis}
We provide a granular visual breakdown of the training dynamics across the 20 independent runs.

\begin{figure}[htbp]
\centerline{\includegraphics[width=\linewidth]{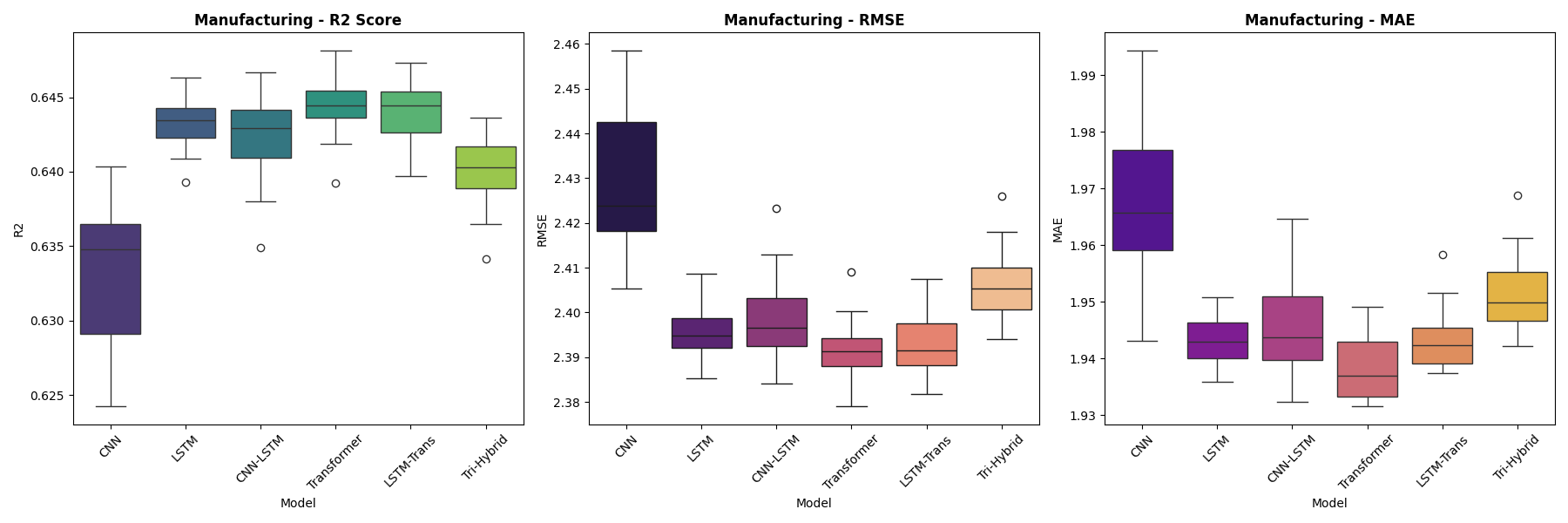}}
\caption{\textbf{Stability Analysis.} The boxplots reveal that LSTM or Attention-based models have significantly lower variance than the Convolutional baseline.}
\label{fig:manuf_box}
\end{figure}

\textbf{1. Stability Analysis (Boxplots):} Fig. \ref{fig:manuf_box} illustrates the distribution of $R^2$ scores, RMSE and MAE. The \textbf{Transformer}, \textbf{LSTM} and \textbf{LSTM-Transformer} exhibit the tightest interquartile ranges (IQR), indicating that these models reliably converge to the global optimum regardless of initialization. In contrast, the \textbf{CNN} displays a wide spread, reflecting its high sensitivity to random seeds \cite{b14}.

\begin{figure}[htbp]
\centerline{\includegraphics[width=\linewidth]{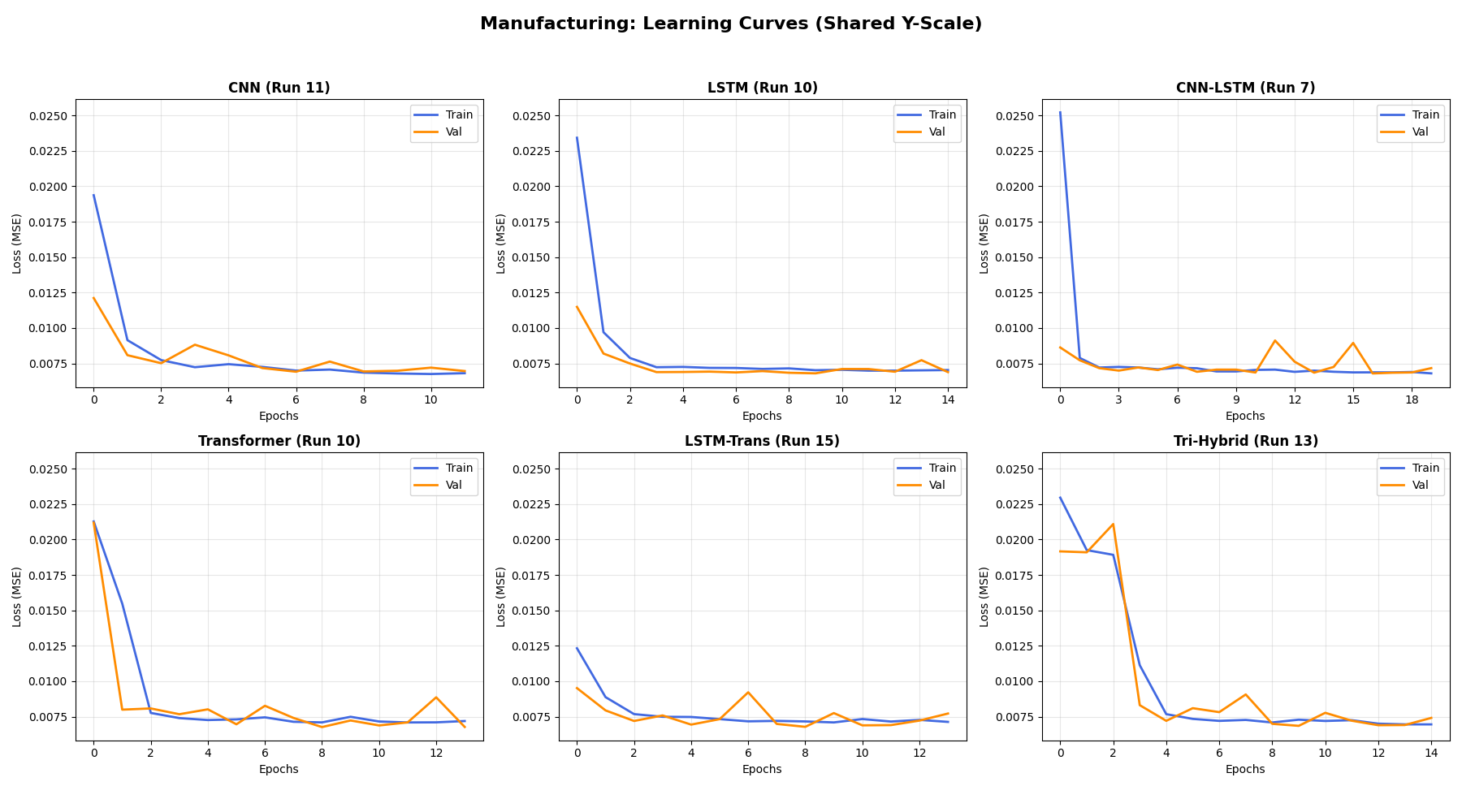}}
\caption{\textbf{Learning Curves.} Comparison of training (Blue) and validation (Orange) loss. All models show stable convergence with minimal overfitting.}
\label{fig:manuf_learning}
\end{figure}

\textbf{2. Convergence Analysis (Learning Curves):} Fig. \ref{fig:manuf_learning} plots the loss trajectories for the best run of each model. All models converged, with validation loss (Orange) closely tracking training loss (Blue). This alignment confirms that Early Stopping and Dropout successfully prevented overfitting, even in the case of complex Tri-Hybrid \cite{b26}.

\begin{figure}[htbp]
\centerline{\includegraphics[width=\linewidth]{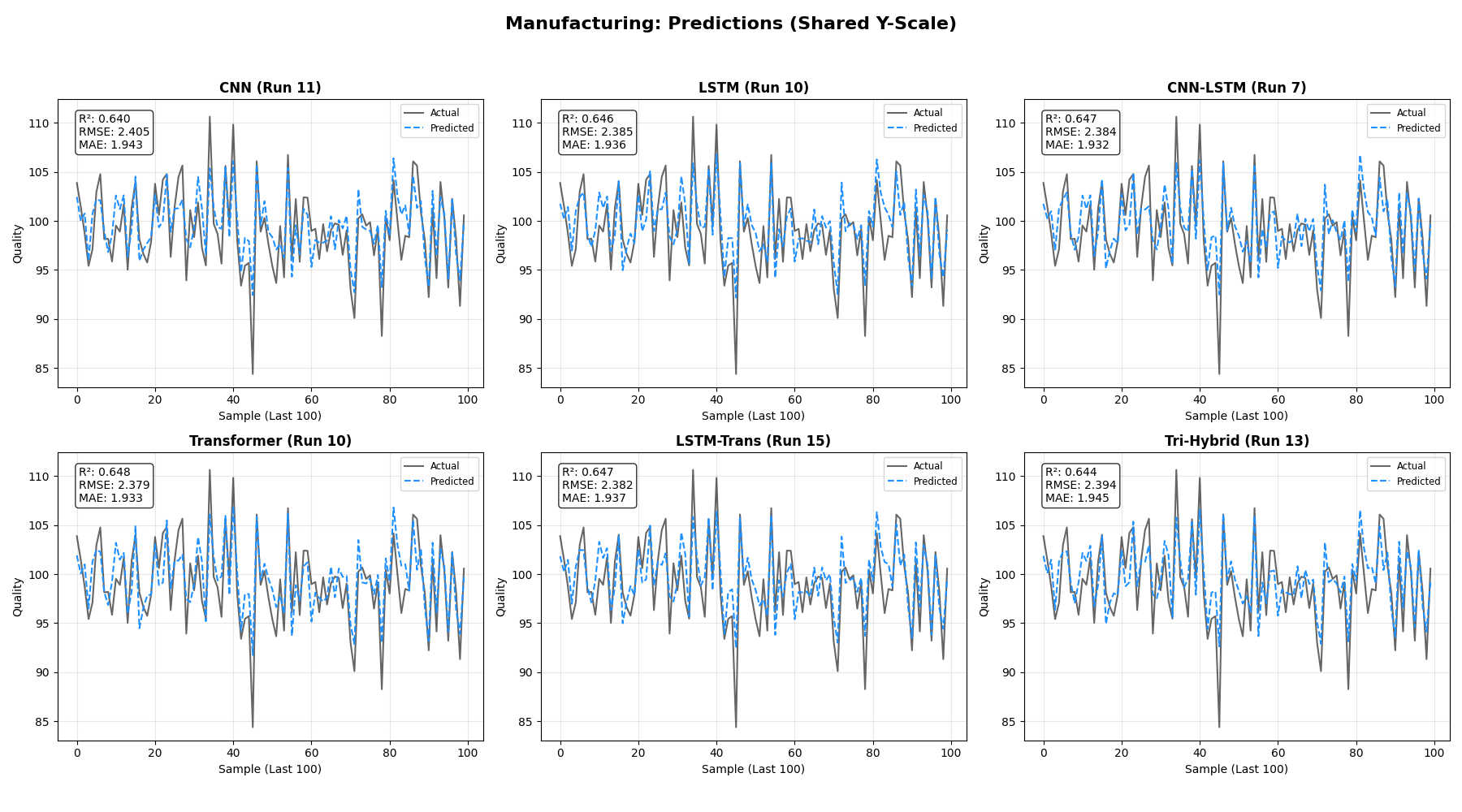}}
\caption{\textbf{Prediction Analysis (Best Run).} While all models track well in their best attempts, the CNN lacks the consistency of memory-based models across repeated trials.}
\label{fig:manuf_pred}
\end{figure}

\textbf{3. Prediction Fit:} Fig. \ref{fig:manuf_pred} displays the \textit{single best run} for each architecture. In this best-case scenario, all models track the ground truth closely ($R^2 \approx 0.64$). However, our stability analysis reveals that the \textbf{CNN} exhibits significantly higher variability across the 20 runs compared to the consistent \textbf{LSTM} and \textbf{Transformer}. It means while the CNN \textit{can} find the pattern, it lacks the memory mechanisms required to do so reliably.

\subsubsection{Discussion: Architectural Analysis}
As shown in Fig. \ref{fig:manuf_box}, the dominance of memory-based models is explained by the physical regime's high inertia. Unlike the memory-less CNN, which relies on local kernels, LSTM and Transformer architectures successfully integrate the accumulated history to predict the process outcome.

\textbf{The Transformer's Advantage:} The Transformer achieves slightly higher precision because its \textit{Self-Attention} mechanism naturally mimics the underlying synthetic regime. Recall that our synthetic target $Y$ is generated using a \textbf{rolling average} (Eq. \ref{eq:manuf_physics}). The Transformer's \textit{Self-Attention} mechanism effectively calculates a weighted average of the entire time window in a single step \cite{b4}. This allows it to mimic the synthetic regime generation process more directly than the LSTM, which must reconstruct the average sequentially through its hidden state.

\subsection{Scenario B: The Chaotic Regime (Augmented IoT)}

In this phase, we assessed the performance of six models on four machine profiles (M001–M004) to see how each Deep Learning architecture handled varying levels of signal stochasticity. Table \ref{tab:iot_full} lists the full performance metrics, based on N=20 runs.

\begin{table}[htbp]
\caption{Model Performance by Machine Type}
\begin{center}
\resizebox{\columnwidth}{!}{%
\begin{tabular}{|l|l|c|c|c|}
\hline
\textbf{Machine} & \textbf{Model} & \textbf{$R^2$ (Mean $\pm$ Std)} & \textbf{RMSE} & \textbf{MAE} \\
\hline
\textbf{M001} & LSTM & 0.450 $\pm$ 0.004 & 8.054 & 7.612 \\
(Chaotic) & Tri-Hybrid & 0.448 $\pm$ 0.006 & 8.066 & 7.614 \\
 & \textbf{LSTM-Trans} & \textbf{0.448 $\pm$ 0.005} & \textbf{8.067} & \textbf{7.591} \\
 & CNN-LSTM & 0.446 $\pm$ 0.006 & 8.083 & 7.615 \\
 & Transformer & 0.443 $\pm$ 0.005 & 8.102 & 7.653 \\
 & ResNet & 0.437 $\pm$ 0.008 & 8.143 & 7.679 \\
\hline
\textbf{M002} & LSTM & 0.509 $\pm$ 0.005 & 7.541 & 6.875 \\
(Inertial) & \textbf{LSTM-Trans} & \textbf{0.508 $\pm$ 0.005} & \textbf{7.553} & \textbf{6.876} \\
 & CNN-LSTM & 0.506 $\pm$ 0.004 & 7.569 & 6.896 \\
 & Tri-Hybrid & 0.505 $\pm$ 0.005 & 7.577 & 6.912 \\
 & Transformer & 0.499 $\pm$ 0.005 & 7.616 & 6.969 \\
 & ResNet & 0.496 $\pm$ 0.007 & 7.639 & 6.988 \\
\hline
\textbf{M003} & \textbf{LSTM-Trans} & \textbf{0.445 $\pm$ 0.008} & \textbf{7.712} & \textbf{7.079} \\
(Mixed) & Tri-Hybrid & 0.443 $\pm$ 0.008 & 7.725 & 7.120 \\
 & CNN-LSTM & 0.438 $\pm$ 0.010 & 7.758 & 7.103 \\
 & LSTM & 0.436 $\pm$ 0.007 & 7.774 & 7.201 \\
 & Transformer & 0.436 $\pm$ 0.009 & 7.778 & 7.198 \\
 & ResNet & 0.428 $\pm$ 0.004 & 7.831 & 7.258 \\
\hline
\textbf{M004} & \textbf{LSTM-Trans} & \textbf{0.485 $\pm$ 0.009} & \textbf{7.867} & \textbf{7.309} \\
(Mixed) & Tri-Hybrid & 0.481 $\pm$ 0.006 & 7.899 & 7.395 \\
 & LSTM & 0.478 $\pm$ 0.005 & 7.924 & 7.464 \\
 & CNN-LSTM & 0.478 $\pm$ 0.007 & 7.925 & 7.416 \\
 & Transformer & 0.477 $\pm$ 0.006 & 7.934 & 7.479 \\
 & ResNet & 0.470 $\pm$ 0.006 & 7.983 & 7.520 \\
\hline
\end{tabular}%
}
\label{tab:iot_full}
\end{center}
\end{table}

\subsubsection{Conclusion on Predictive Power}
The results are split between four machine types.
\begin{itemize}
    \item \textbf{M001 (Chaotic):} The \textbf{LSTM-Transformer} achieves the lowest MAE ($7.591$), tracking the underlying machine degradation trend better than the standard LSTM ($7.612$), when noise levels are high.
    \item \textbf{M002 (Inertial):} While the standard \textbf{LSTM} performs slightly better ($R^2=0.509$), the attention-based architectures remain highly competitive ($R^2=0.508$ for LSTM-Trans).
    \item \textbf{M003 \& M004 (Mixed):} The \textbf{LSTM-Transformer} outperforms the complex Tri-Hybrid in both mixed regimes ($R^2=0.445$ and $0.485$). This proves that a hybrid design (Filter + Attention) captures signals from mixed simulations more effectively than simply increasing model complexity (i.e., parameter-heavy Tri-Hybrid).
\end{itemize}

\subsubsection{Visual Analysis}
\begin{figure*}[htbp]
    \centering
    \begin{minipage}{0.48\textwidth}
        \centering
        \includegraphics[width=\linewidth]{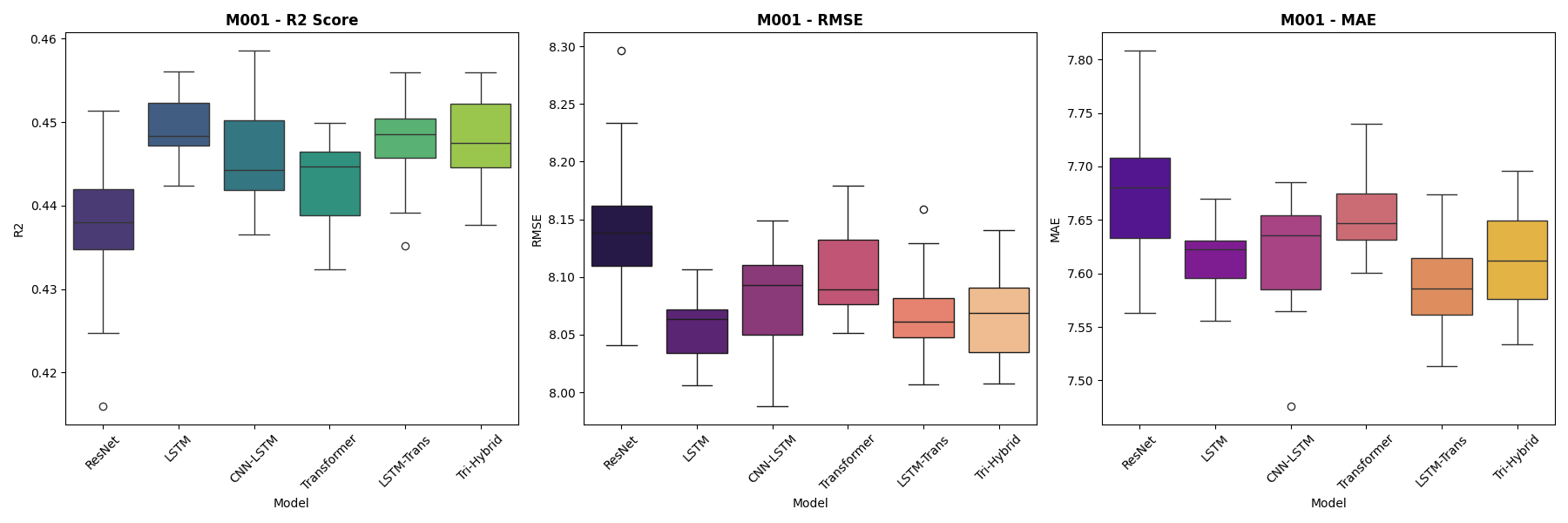}
        \par\vspace{2pt}
        \footnotesize (a) Stability Analysis
    \end{minipage}\hfill
    \begin{minipage}{0.48\textwidth}
        \centering
        \includegraphics[width=\linewidth]{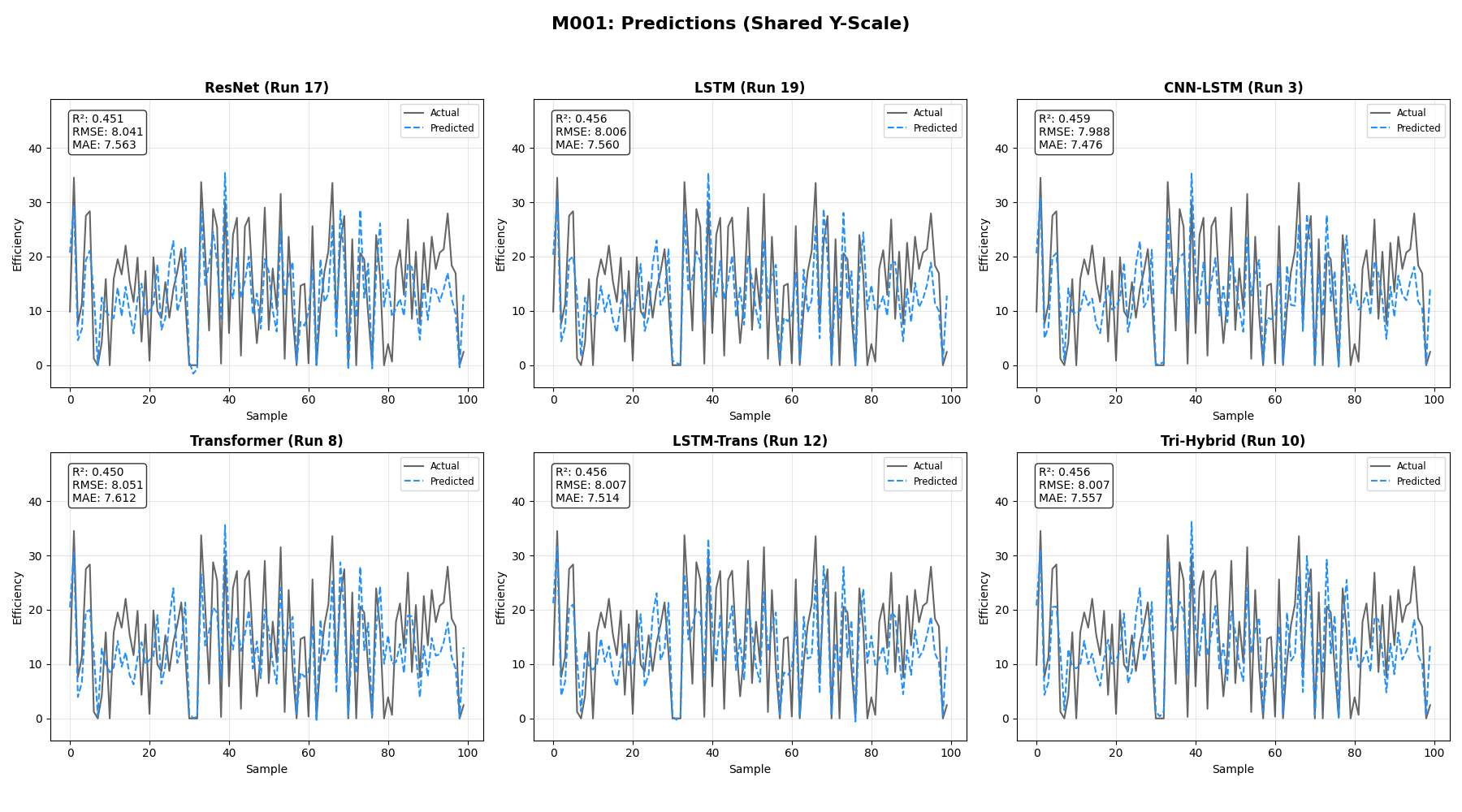}
        \par\vspace{2pt}
        \footnotesize (b) Prediction Analysis (Best Run)
    \end{minipage}
    \caption{\textbf{Analysis of Chaotic Regime (M001).} (a) Only LSTM-based models maintain tight variance; others show significant instability. (b) In the ``Best Run" scenario, performance is visually identical, highlighting the need for stability metrics.}
    \label{fig:m001_composite}
\end{figure*}

We focus on Machine M001 to visualize the challenge of chaotic regimes. Fig. \ref{fig:m001_composite} displays the stability distribution (Boxplot) and the time-series prediction (Prediction Plot).

\textbf{1. Stability Analysis (Boxplots):} The Boxplots in Fig. \ref{fig:m001_composite}(a) reveal a clear stability divide. With the exception of the \textbf{LSTM} and \textbf{LSTM-Transformer}, all other architectures (ResNet, CNN-LSTM, Transformer, Tri-Hybrid) exhibit large interquartile ranges. The \textbf{LSTM-Transformer} maintains a compact variance, providing reliable convergence even when the signal is stochastic \cite{b8}.

\textbf{2. Prediction Fit:} The Prediction Plots in Fig. \ref{fig:m001_composite}(b) visualize the \textit{single best run}. In this optimized scenario, visual differences vanish; all models achieve similar scores ($R^2 \approx 0.45$) and track the volatility. This parity confirms that the primary challenge in industrial AI is not achieving capacity, but achieving \textit{reliability}, effectively making the stability metrics in Fig. \ref{fig:m001_composite}(a) the deciding factor for deployment.

\subsubsection{Discussion: Architectural Analysis}
A critical finding of this study is the counter-intuitive success of the LSTM-Transformer on the chaotic M001 dataset. Given that M001 exhibits weak temporal dependence (Fig. \ref{fig:acf_comparison}), one might expect memory-based architectures to fail. However, our results suggest that the LSTM layer is not functioning as a long-term memory unit, but rather as a \textbf{Sequential Noise Filter} \cite{b9}.

\begin{itemize}
    \item \textbf{Transformer Failure Mode:} The pure Transformer's global attention mechanism considers all time steps simultaneously. In the high-noise regime of Dataset 2 ($\sigma=2.0$), the model struggles to distinguish between the underlying physical signal and Gaussian noise, leading to overfitting and high variance \cite{b6}.
    \item \textbf{LSTM Filtering Effect:} By processing the input sequentially, the LSTM layer decides what information should be kept and what should be dropped. It effectively acts as a learnable low-pass filter, suppressing high-frequency stochasticity and generating stable hidden states ($h_t$). When these ``denoised'' representations are passed to the Transformer block, the attention mechanism can then safely identify non-linear relationships without being disrupted by raw sensor noise \cite{b7}.
\end{itemize}

A limitation of this analysis is that the injected noise is Gaussian, which is symmetric and stationary. Real industrial noise often includes sensor drift, intermittent dropouts, and heavy-tailed distributions. Whether the LSTM filtering effect generalizes to such non-stationary disturbances remains an open question for future work.

Nevertheless, this finding suggests that in noisy industrial environments, the value of Recurrent Neural Networks lies not just in their ability to remember the distant past, but also in their ability to stabilize the immediate present.

\subsection{Scenario C: The Global Test}

In the final phase, we evaluated the models on the ``Global'' dataset, which concatenates data from all four machines. This tests the architecture's capacity to learn a generalized representation of tool wear across distinct machine profiles \cite{b11}.

\subsubsection{Conclusion on Predictive Power}
Table \ref{tab:global_results} presents the aggregated performance metrics.
\begin{itemize}
    \item \textbf{The Generalization Leader:} The \textbf{LSTM-Transformer} achieves the highest Mean $R^2$ ($0.472$) and the lowest error rates (RMSE: $7.823$, MAE: $7.323$).
    \item \textbf{Lagging Baselines:} The pure \textbf{ResNet} and \textbf{Transformer} models fell to the bottom of the ranking, confirming that single-domain architectures struggle to adapt to the mixed simulated regimes of a global fleet.
\end{itemize}

\begin{table}[htbp]
\caption{Global Dataset Results}
\begin{center}
\begin{tabular}{|l|c|c|c|}
\hline
\textbf{Model} & \textbf{$R^2$ Score (Mean $\pm$ Std)} & \textbf{RMSE} & \textbf{MAE} \\
\hline
\textbf{LSTM-Trans} & \textbf{0.472 $\pm$ 0.004} & \textbf{7.823} & \textbf{7.323} \\
CNN-LSTM & 0.472 $\pm$ 0.003 & 7.825 & 7.330 \\
Tri-Hybrid & 0.472 $\pm$ 0.004 & 7.828 & 7.334 \\
LSTM & 0.471 $\pm$ 0.002 & 7.833 & 7.347 \\
Transformer & 0.471 $\pm$ 0.004 & 7.837 & 7.350 \\
ResNet & 0.470 $\pm$ 0.004 & 7.843 & 7.345 \\
\hline
\end{tabular}
\label{tab:global_results}
\end{center}
\end{table}

\subsubsection{Visual Analysis}
\begin{figure*}[htbp]
    \centering
    \begin{minipage}{0.48\textwidth}
        \centering
        \includegraphics[width=\linewidth]{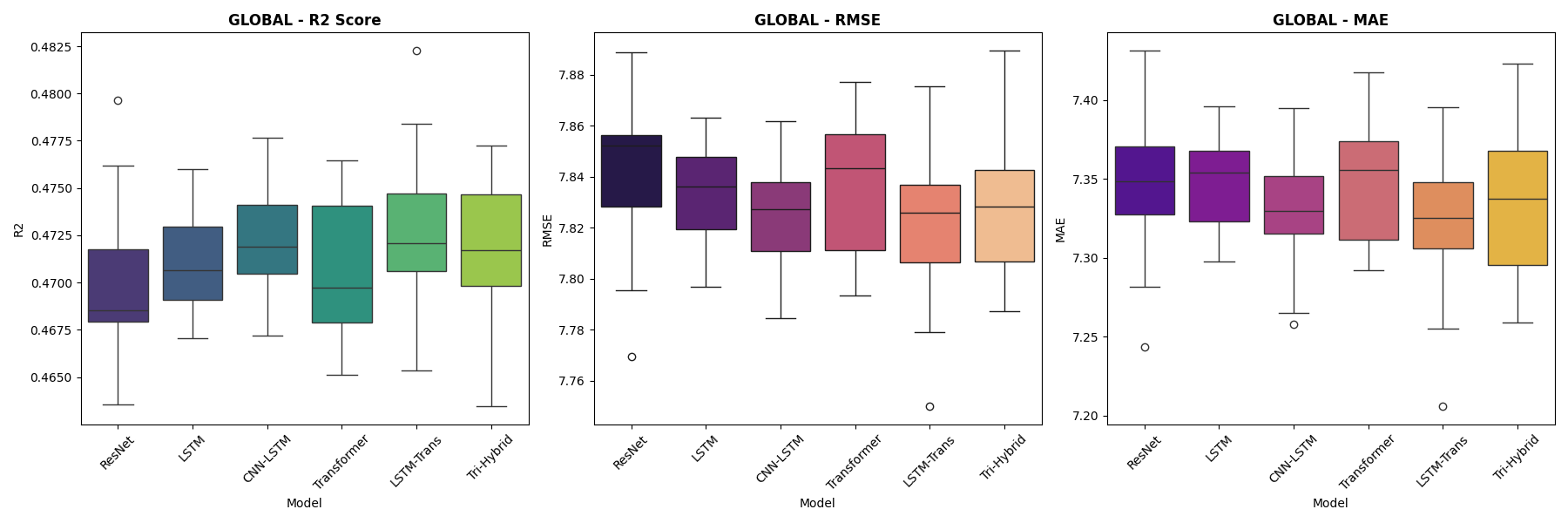}
        \par\vspace{2pt}
        \footnotesize (a) Stability Analysis
    \end{minipage}\hfill
    \begin{minipage}{0.48\textwidth}
        \centering
        \includegraphics[width=\linewidth]{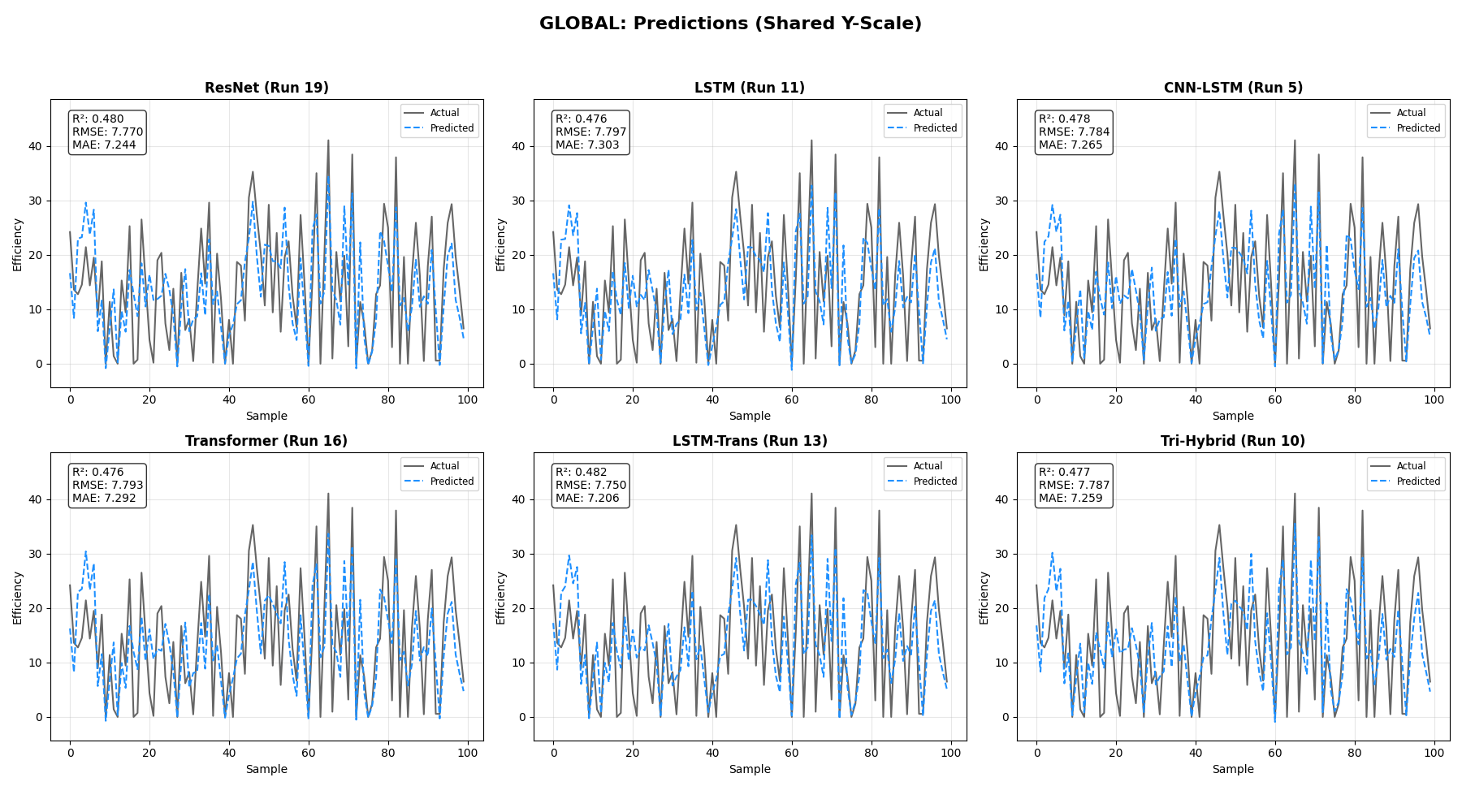}
        \par\vspace{2pt}
        \footnotesize (b) Prediction Analysis (Best Run)
    \end{minipage}
    \caption{\textbf{Analysis of Global Test.} (a) The mixed dataset increases variance for all models, but the LSTM-Transformer remains the most stable. (b) The ``Best Run" plots are virtually identical, confirming that capacity is not the bottleneck—consistency is.}
    \label{fig:global_analysis}
\end{figure*}

Fig. \ref{fig:global_analysis} illustrates the challenge of the ``Global" experiment, where models must learn conflicting machine profiles (Stable vs. Chaotic) simultaneously.

\textbf{1. Stability Analysis (Boxplots):} Fig. \ref{fig:global_analysis}(a) reveals the cost of diversity. Mixing distinct machine profiles causes the variance to increase across \textit{all} architectures compared to the single-machine experiments in \textit{Scenario B}. However, the \textbf{LSTM-Transformer} (Green) proves to be the most resilient. While the \textbf{Tri-Hybrid} and \textbf{Transformer} exhibit wide interquartile ranges, the LSTM-Transformer maintains the lowest median error (RMSE/MAE) and the most controlled variance.

\textbf{2. Prediction Fit:} Fig. \ref{fig:global_analysis}(b) displays the single best run. In this idealized scenario, every model demonstrates sufficient capacity to track the trend ($R^2 \approx 0.48$). Because the visual performance is almost identical in the best case, the stability metrics in Fig. \ref{fig:global_analysis} (a) become the reliable criteria for selection.

\subsubsection{Discussion: Architectural Analysis}
The Global experiment confirms the \textbf{LSTM-Transformer} as the most reliable architecture for mixed regimes. Its success comes from a hybrid design that handles conflicting signal regimes. The \textbf{LSTM layer} acts as a sequential filter, dampening the high-frequency noise from chaotic machines (e.g., M001). Simultaneously, the \textbf{Transformer layer} uses these stabilized states to capture the slow-moving, long-range degradation trends of inertial machines (e.g., M002) \cite{b7}. This distinct separation of duties allows the model to adapt to the mixed dataset without the instability seen in pure Transformers or the extra complexity of the Tri-Hybrid model.

\section{Conclusion}
We presented a stability-focused evaluation of six Deep Learning architectures for predictive maintenance. Instead of reporting a single best-case result, we ran more than 700 independent trials in both stable and chaotic regimes to more accurately assess model reliability.  

Our analysis offers three practical takeaways for the industrial AI community:

\begin{enumerate}
    \item Pure Transformers struggle when data gets chaotic, often confusing noise for signal \cite{b6}. We identified the \textbf{LSTM-Transformer} as the definitive solution. By using the LSTM layer to ``clean” the noise step-by-step, it protects the Attention mechanism, resulting in the most stable performance across diverse machines.
    
    \item Our stress tests revealed that bigger isn’t better. The complex \textbf{Tri-Hybrid} model failed to beat the simpler LSTM-Transformer and was often less stable. This confirms that smart design, specifically combining a filter with an attention block, is superior to blindly stacking layers \cite{b18}.
    
    \item The \textbf{Autocorrelation Function (ACF)} can serve as a practical guide for model architecture choice. High-inertia physical systems need an Attention layer for capturing longer-term patterns \cite{b4}. On the contrary, in more chaotic settings, recurrent layers can be a better fit for handling short-term volatility \cite{b3}.
\end{enumerate}

In summary, while ``all-attention'' models lead the most current research, the stochastic reality of the factory floor demands resilience. The LSTM-Transformer stands out as the most reliable engine for scalable, regime-agnostic predictive maintenance modeling.

\end{document}